\documentclass[letterpaper, 10 pt, conference]{ieeeconf}  

\IEEEoverridecommandlockouts                              

\usepackage{stfloats}
\usepackage{float} 

\usepackage{booktabs} 
\usepackage{tabularx} 
\title{\LARGE \bf
Stakeholder Insights for Designing In-Home Social Robots for Dementia Disorientation Detection and Caregiver-Aware Intervention
}

\author{Emmanuel Akinrintoyo and Nicole Salomons 
\thanks{Emmanuel Akinrintoyo and Nicole Salomons are with  Imperial College London
        {\tt\small, e.akinrintoyo23@imperial.ac.uk,
n.salomons@imperial.ac.uk}}%
        }%

\usepackage{graphicx}
\usepackage{dirtytalk}
\usepackage{booktabs} 
\usepackage{amsmath}

\begin{document}

\maketitle
\thispagestyle{empty}
\pagestyle{empty}

\begin{abstract}
Dementia disorientation detection and intervention remain under-examined as socio-technical challenges for socially assistive robots (SARs). We conducted 14 semi-structured interviews with dementia caregivers and practitioners (DCPs) to investigate how disorientation is experienced, recognised, and managed in everyday life. The findings reveal that disorientation is recurrent and fluctuating. It often emerges through behavioural cues such as repeated questioning, inappropriate activity timing and disrupted daily routines. Caregivers described orientation as emotionally charged, and direct correction may increase distress. The DCPs were generally receptive to robotic assistance when framed as supportive rather than corrective. Based on the insights, we identify essential design implications for SARs that provide context-aware orientation support, integrate into daily routines and support caregivers through timely escalation. 
\end{abstract}

%


\section{Introduction}

Dementia is one of the foremost health challenges in the twenty-first century~\cite{barros2020dementia}. It is estimated that over 152 million older adults worldwide will have dementia by 2050~\cite{nichols2022estimation}. Dementia causes a progressive decline in the cognitive capabilities of older persons~\cite{boller1998history, mendez2003dementia} such that persons living with dementia (PwDs) experience disorientation with time, place and person~\cite{chiu2018reality}. 

Disorientation is a particularly distressing consequence of dementia~\cite{amaefule2020effect, better2023alzheimer}. It limits the ability of PwDs to live independently~\cite{cipriani2020daily}. It disrupts their daily activities. PwDs forget their schedule for daily activities such as medication intake, feeding and hygiene~\cite{jefferson2006cognitive}. Disorientation is also accompanied by increased levels of anxiety, stress and agitation~\cite{kales2015assessment}. The combined effects of disorientation result in a lower quality of life for PwDs~\cite{banerjee2006quality}. Additionally, it increases the caregiver burden and the risk of institutionalisation~\cite{gaugler2009predictors, brodaty2009family}.

While dementia is often considered a cognitive deficit, disorientation emerges in daily life as a dynamic interactional challenge rather than a single observable event~\cite{hamilton1994conversations, kindell2017everyday, schaat2020realtime}. It unfolds through changes in behaviour, routines and interaction patterns~\cite{yang2025multimodal}. This can include repeated moments in which a PwD seeks clarification, prepares for activities at inappropriate times, or attempts actions that no longer align with their context. These situations create the need for caregivers to recognise and respond to disorientation. 



Existing assistive technologies provide limited support for disorientation. The current approaches include reminder systems~\cite{mcgoldrick2021mindmate, akinrintoyo2025user}, wayfinding aids~\cite{marquardt2011wayfinding, kwan2020use}, and fall detection~\cite{bharucha2009intelligent, dada2024intelligent, gettel2021dementia}. These technologies often function as passive information systems that provide prompts or alerts rather than recognising and responding to disorientation as it emerges. Hence, caregivers are mainly responsible for recognising and responding to disorientation in real time.


The embodiment and interactivity of robots offer a promising solution to address this gap. Rather than simply providing information, SARs offer the potential to detect behavioural cues of disorientation, such as repeated questioning or a person initiating activities at inappropriate times. They can also provide contextual support and remind a user of the current time or situation in less confrontational ways than human correction. Rather than replacing caregivers, a SAR can provide routine prompts and orientation cues to reduce the frequency of caregiver intervention. However, limited work has explored how SARs can support disorientation detection and intervention in everyday life. 


In this work, we investigate disorientation detection and intervention in the homes of PwDs as an interactional and socio-technical challenge for SARs. We conducted semi-structured interviews with 14 DCPs. 
The interviews explored aspects of disorientation and the lived experiences of DCPs. Based on the findings of the interviews with DCPs, we show that an effective system should integrate into daily routines, provide simple orientation support, and remain sensitive to PwDs' lived experiences. 

\section{BACKGROUND}
\subsection{Disorientation}
Disorientation is a core cognitive symptom of dementia~\cite{berry2014minimizing, chiu2018reality}. It affects an individual’s awareness of time, place, and person~\cite{chiu2018reality}. 
Temporal disorientation can manifest as confusion about the time of day or the sequence of events~\cite{johansson2001experience}. Spatial disorientation can involve difficulty navigating familiar spaces or identifying one's own home~\cite{liu1991spatial}. Person disorientation can cause challenges with recognition of caregivers, family members or friends.

Notably, disorientation is not a binary state~\cite{spector2000reality, chiu2018reality}. It can fluctuate daily for each PwD and is context-dependent. This implies that a PwD may appear oriented in structured or familiar situations yet become disoriented under stress, increased cognitive load or environmental change. This variability is a challenge for clinical assessment and technological support. Hence, it points to the need for detection and intervention that is continuous and context-aware. 

\subsection{Technology-Based Disorientation Detection in the Home}

Recent work has explored technology-enabled environments to monitor changes in behavioural indicators of disorientation~\cite{cook2012casas, rashidi2012survey}. Smart home systems can be equipped with motion, door and environmental sensors that monitor the deviations in daily living activities. For example, Dawadi et al.~\cite{dawadi2013automated} used machine learning and environmental sensors in a smart home to assess activity performance and monitor behavioural signals. Evaluated with 179 participants, including PwDs, the system showed strong correlations between automated activity-quality scores and clinical assessments. 

In dementia care, IoT-based assistive technologies have mostly focused on wandering behaviour and spatial disorientation. Sensor systems such as geofencing devices, wearable trackers, and location monitoring devices are often used to identify wandering behaviour and notify caregivers. 
Lin et al.~\cite{lin2015disorientation} used historical GPS data to model an older person’s regular movement patterns to detect disorientation by identifying abnormal trajectories. It had a detection rate of approximately $95\%$ with a false positive rate below $3\%$. However, the solutions mostly focus on other forms of disorientation that occur in routine activities at home.
 
Research on digital biomarkers highlights the potential of mobile and wearable technologies to detect early behavioural and physiological signs of disorientation. Sensors in smartphones and wearables can capture signals such as gait and speech~\cite{kourtis2019digital}. For example, Schaat et al.~\cite{schaat2020real} used wearable accelerometer data to detect spatial disorientation in individuals with cognitive impairment (AUC of $0.75$). 
However, these approaches mainly focus on detection and offer limited real-time support for PwDs in everyday settings.

\subsection{Social Robots and Responsible Intervention in Dementia}
Previous work on SARs in dementia care focused on therapeutic, emotional, and cognitive outcomes~\cite{joranson2016change, moyle2017use, akinrintoyo2026, akinrintoyo2026_2}. They sought to reduce episodes of symptoms (e.g., agitation and depression). Majority utilised the pet-type PARO robot~\cite{joranson2015effects, joranson2016change, liang2017pilot, moyle2017use, pu2020effect}. They found positive effects on mood, engagement, agitation, and depressive symptoms in residential and clinical environments. However, the interventions were often short-term and symptom-focused, rather than providing continuous, context-aware support.

Other studies explored humanoid robots (e.g., Sil-bot~\cite{park2021humanoid}, and Eva~\cite{cruz2020social}). They recorded improvements in memory, executive function, and depressive symptoms such as agitation and delusions among older adults with mild cognitive impairment. Improvements in daily activities and cooperation, such as participation in activities and social engagement, were noted~\cite{cruz2020social}. Notably, cognitive function was boosted~\cite{park2021humanoid}. Yet, these interventions were generally structured, activity-based and delivered within controlled settings.

Despite these advances, limited consideration has been given to how SARs might detect and respond to disorientation as it unfolds in daily life. Existing systems have primarily focused on structured therapeutic activities. They are rarely designed to recognise behavioural indicators of disorientation, such as repeated questioning or engaging in activities at unusual times. Little is known about how robots could support PwDs during episodes of confusion. 
To address this gap, this work seeks to investigate the main challenges and design considerations for providing robot-based disorientation intervention and support to PwDs. Through consultations with DCPs, we explore how SARs can aid PwDs by recognising behavioural cues of confusion in everyday living activities and routines. We identify opportunities and challenges for providing timely, context-sensitive support.

\section{INTERVIEWS}
We conducted 14 semi-structured interviews with DCPs. This comprised of DCPs who support PwDs daily, such as professional caregivers (PC1–PC3) and family caregivers (FC1–FC4), an occupational therapist (OT1), dementia nurse practitioners (DNP1–DNP2), well-being and technology leads (WTL1–WTL2), and mental health practitioners (MHP1–MHP2). The participants provided complementary perspectives across caregiving, therapeutic and service design roles. Microsoft Teams was used to conduct one-hour interviews. Participants were prompted to discuss their lived experiences with disorientation in caring for and supporting PwDs daily. It included discussions on how frequently they observed disorientation in PwDs daily, the most common type of disorientation they encountered, the challenges they encountered in supporting PwDs in a disorientated state, and how a SAR can support PwDs to such a state. Participants also provided diverse insights on how interventions are delivered, and the challenges associated with timing, communication and emotional impact. Each participant was compensated with 15 pounds for their time. The study had ethical approval (ethics no: 6871649).

\subsection{Results}
We conducted a qualitative content analysis of interview responses related to disorientation, its detection, and intervention. The interview transcripts were transcribed and systematically coded to identify recurring categories and patterns in the stakeholders' accounts. Subsequently, an iterative process was used to refine and group the categories by making comparisons across the transcripts. The resulting themes are presented below:


\subsubsection{Theme 1: Disorientation is a Recurrent and Fluctuating Experience} 
All participants consistently described that disorientation is a common experience for PwDs. Participants discussed that disorientation is often not a single, clearly defined moment for PwD. Caregivers often recognised it through repeated behaviours rather than immediate cognitive errors. PC1 explained that \textit{\say{there would be some residents who would be disoriented most of the time}}. Caregivers often inferred disorientation when attempts at orientation failed or when confusion occurred shortly after an orientation. Participants also described disorientation as fluctuating rather than constant. Instead, it could recur throughout the day, even within short intervals. FC1 noted that \textit{\say{they look at it [dementia memory clock] sometimes now, but then they forget right after again}}. 

Eleven participants reported temporal disorientation as the most common form. Others emphasised place disorientation. PC3 described that \textit{\say{they might experience temporal disorientation and not know what day it is or what time it is}}. Eight participants noted that it was particularly a problem at night because staff numbers are low and the family is absent. WTL1 described that \textit{\say{they might get from bed to the front door without anyone stopping them}}. 

In addition, disorientation was regarded as emerging within everyday routines. It was identified through indicators such as being awake at night, waiting for meals at inappropriate times, repetitive phone calls, missing activities, and increased agitation during late afternoon and evening (referred to as sundowning). Other signs include temporal slips in conversation, changes in the use of language and narrative time shifts. FC2 described instances when her family was \textit{\say{thinking it’s like five years earlier}}.

Participants suggested some potential robotic interventions. WTL1 described that \textit{\say{the robot could interrupt autopilot thinking at night, that would help}} (WTL1). PC3 advised that \textit{\say{a robot could say: it’s dark outside, it’s night, it’s not time to get up yet}}.

\noindent\textbf{Design Implication:} SARs should treat disorientation as a dynamic and recurring phenomenon revealed through patterns of behaviour over time. They should interpret behavioural cues based on routine activities, temporal context, and deviations from daily patterns rather than responding to isolated signals. This may support more effective and contextually appropriate orientation for PwDs.

\subsubsection{Theme 2: Aiding Orientation Is Challenging and Requires Careful Communication}
Six participants noted their motivation to use technology to support PwDs is because of how overwhelming disorientation is. They described that aiding orientation was often described as a complex and often fragile interactional process. This was strongly present in the accounts of the four family caregivers. They described the emotional strain and exhaustion that it creates. FC3 stated that \textit{\say{I just can’t do it anymore, nights are the worst}}. FC3 described that \textit{\say{there aren’t enough people to do this care, something else has to help}}. Hence, a sustained interactional labour is required to support PwDs.

Three family caregivers highlighted the emotional burden of repeated reorientation. Five professionals emphasised the emotional risks involving correction itself. Repeated or poorly timed reorientation efforts were often overwhelming for some PwDs. It sometimes led to increased agitation or difficult behaviour. PC1 described that \textit{\say{trying to reorient them back can cause them to become overwhelmed, and that can then cause their behaviour to become challenging}}. 

Reorientation strategies were often initially effective. However, their impact was frequently short-lived. Hence, it required repeated effort with limited lasting benefit. One participant noted that \textit{\say{earlier on we had a lot of success with reorientating them}} (FC1). FC1 described that \textit{\say{it doesn't matter whether we actually tell him [the time of day], (...) that doesn't work anymore}}.

\noindent\textbf{Design Implication:} Orientation support should prioritise emotional well-being with informational accuracy. SARs should avoid corrective interactions with PwDs. Instead, they should offer adaptive, supportive prompts. Such interactions should reduce the risk of distress or agitation while also reducing caregiver burden.

\subsubsection{Theme 3: Robotic Intervention Is Acceptable When It is Supportive and Context-Aware}
Twelve of the fourteen participants, including all family carers, were generally open to robotic intervention when it was framed as a supportive aid for disorientation. Acceptance was closely linked to their expectations of how the robot would behave. Caregivers considered the social robot's intervention potentially helpful in low-risk, routine situations where reminders or its presence could supplement their care provision. WTL1 noted that \textit{\say{I think if it could intervene that would be really helpful}}. 

Nine participants described specific contexts in which robotic support would be appropriate. This can include the provision of reminders in familiar domestic spaces. A caregiver suggested that \textit{\say{it would be great if it could sit in the kitchen and just remind them}} (FC1). This highlights the expectations of caregivers. The social robot should provide simple assistance that is not intrusive, but be aware of the personal preferences of a PwD. WTL1 noted that \textit{\say{the robot should know if the person likes being told or asked}}. This shows a link between acceptance and the perceived scope of the robot's actions. 

PwDs in the early stages of dementia were described as engaging well with simple and familiar technologies. This included tablets and dementia memory clocks. They also use non-digital aids such as large calendar displays and whiteboards for daily structure and orientation. These tools were valued for supporting autonomy, independence and self-orientation. PC2 stated that PwDs could \textit{\say{if I don’t put a clock in, they don’t know. If I put the clock in, they can work it out themselves}}. However, the ability to learn and adapt to new technologies declines as dementia progresses. 

The participants described that concerns around privacy, trust and recording would vary by person. PwD's acceptance of robotic sensing, particularly audio input, would depend on their individual differences and personal preferences. DNP2 mentioned that \textit{\say{yes it depends on the person}}. Hence, trust in robotic systems cannot be assumed. However, they noted that such preferences are influenced by previous familiarity with technology. DNP1 stated that \textit{\say{I think this is dependent on people’s kind of background}}. 

\noindent\textbf{Design Implication:} The acceptance of robotic support is shaped by perceived usefulness, personalisation, and trust. SARs should provide context-aware assistance that aligns with a PwD's preferences and routines. Moreover, users and caregivers should have control over the sensing and privacy settings of the robots. Framing the interventions as supportive rather than corrective may further enhance acceptance.

\subsubsection{Theme 4: Non-Intervention and Emotional De-escalation as Deliberate Care Strategies}
Oftentimes, caregivers did not take an active correction or reorientation step when they detected disorientation. Instead, they described that their decisions were based on previous experiences and the awareness of a mistimed correction. This occurred when a PwD reverted quickly to confusion. In such instances, restraint was the approach to reducing and avoiding frustration or stress. PC3 noted that \textit{\say{sometimes we tell the truth, sometimes we don’t, to reduce stress}}. 

The emotional consequences of human-led corrections influence corrections made to PwDs. PC2 discussed that \textit{\say{we don’t correct them sometimes}} because direct corrections could provoke negative reactions. PC2 added that \textit{\say{they get a bit angry with you because that’s not their reality}}. Human corrections can be experienced as confrontational or invalidating by PwDs. PC2 suggested that technology can be less emotionally charged in the support it provides: \textit{\say{it’s easier to hear it from a computer than from ourselves, because we irritate them by correcting them all the time}}. 

\noindent\textbf{Design Implication:} SARs should be capable of selective intervention. The robots should not treat every occurrence of disorientation as a problem to be corrected. Instead, they should adapt their responses to the emotional context and potential consequences of intervention. A gentle reassurance, distraction, or non-intervention may be more suitable in certain circumstances rather than direct reorientation.

\subsubsection{Theme 5: Caregiver Notification and Escalation Are Essential}
Ten participants emphasised the need for caregiver involvement when disorientation is detected in some specific instances. This includes situations that involve safety risks or uncertainty. While the robotic interventions were considered beneficial, caregivers did not expect robots to replace human decision-making in higher-risk scenarios. Rather, the DCPs described their preference for SAR systems that can recognise when an escalation to a caregiver is warranted. Notification preferences were described as context-dependent. The preferences should be influenced by factors such as the severity of disorientation and the time of day.

Six caregivers noted that they have a clear preference for timely and direct notification when intervention is needed. For example, FC1 noted that \textit{\say{I think I’d prefer [to be] called because maybe it’s more urgent}}. This reflects situations where robotic support alone may be insufficient and human judgment becomes necessary. FC2 explained that in instances of unsupervised or unmanaged disorientation, PwDs \textit{\say{can do damage to themselves}}. 

The robot's operation must not create added complexity or increase the caregiver's burden. Seven participants suggested that the tolerance of a caregiver for a system strongly influences whether an intervention is adopted, maintained or abandoned. The system should reduce caregiver work. MHP1 noted that \textit{\say{the carer is rather busy}}. Systems that provided support in high-risk periods or reduced the need for constant supervision were regarded as more acceptable. MPH2 reflected that \textit{\say{if the robot was sufficient to get the person to return to bed and not disturb the carer… wouldn’t that be a much better outcome all around}}. This is notable especially if it could \textit{\say{help alleviate sort of care burden and… allow them to facilitate the person living at home for as long as possible}} (MPH2). MPH2 referred to this as the \textit{\say{the goal of most… dementia services}}. 

\noindent\textbf{Design Implication:}
SARs should identify when disorientation exceeds the scope of automated support and notify caregivers accordingly. Thus, escalation mechanisms should be integrated for robots to operate as part of a broader care network. SARs should be considered as a supportive tool that balances autonomy with coordination.


\subsubsection{Theme 6: Integration into daily routines}
Eight participants described that long-established daily routines previously supported orientation for PwDs. The routines reduce the need of PwDs for explicit orientation, such as through verbal instructions. Temporal information of days and times retained meaning when the routines were intact. However, disorientation became a challenge when the routines broke down. WTL1 noted that \textit{\say{Once routine goes, everything else starts to unravel very quickly}}. OT1 and MPH2 linked this to time confusion. MPH1 stated that \textit{\say{They don’t know it’s Tuesday any more because Tuesday doesn’t mean anything without the routine attached to it}}. Two of the four family caregivers also echoed this. FC3 reflected that \textit{\say{She used to have structure for everything—washing days, meal times—once that disappeared, the confusion really increased}}.

\noindent\textbf{Design Implication:}
Robots should be designed to operate within the routine structures that organise daily life for PwDs. Orientation should be embedded within routine activities. It should be delivered in relation to familiar behaviour patterns. Detecting disruptions to daily routines may provide opportunities for early intervention when disorientation emerges.


\section{Discussion}

\subsection{Disorientation as an Interaction Problem}
The findings suggest that disorientation is rarely identified through cognitive errors alone. Caregivers rarely relied on one cognitive error to recognise disorientation. 
SARs should interpret disorientation as a behavioural pattern that unfolds over time.


\subsection{Robots as Supportive Interaction Partners}
Caregivers emphasised that orientation support is emotionally sensitive. Direct correction can lead to frustration or agitation, especially when it challenges a PwD's perceived reality. They chose not to intervene in some situations. 
SARs should provide orientation information as suggestions, prompts, or conversational cues rather than commands. 


\subsection{Long-Term Integration into Routine}
Established routines were described as essential anchors for orientation, and their breakdown often intensified disorientation. Participants described that assistive technologies were most effective when introduced early and aligned with familiar practices. SARs should be embedded gradually within the daily routines of PwDs rather than as a corrective tool. 

\section{Limitations}
We acknowledge that the perspective of the caregivers and dementia professionals may differ from the lived experiences of PwDs themselves. Future work will engage with PwDs before a long-term deployment in their homes.

\bibliographystyle{IEEEtran}
\bibliography{IEEEabrv,bibliography}


\end{document}